\RequirePackage{fix-cm}
\documentclass[twocolumn]{svjour3}          
\smartqed  
\usepackage{graphicx}
\usepackage{epsfig}
\usepackage{epstopdf}
\usepackage{subfigure}
\usepackage{amssymb}
\usepackage{amsmath}
\usepackage{bm}
\usepackage{multirow}
\usepackage{booktabs} 
\usepackage{hyperref}
\usepackage{adjustbox}
\newcommand{\eat}[1]{}
\usepackage{blindtext}
\usepackage[ruled]{algorithm2e}

\begin{document}

\title{Learning in High-Dimensional Multimedia Data: The State of the Art
\eat{\thanks{Grants or other notes
about the article that should go on the front page should be
placed here. General acknowledgments should be placed at the end of the article.}
}}


\author{Lianli Gao         \and
        Jingkuan Song$^*$  \and
        Xingyi Liu	\and
        Junming Shao \and
        Jiajun Liu \and
        Jie Shao
}


\institute{Lianli Gao, Junming Shao, Jie Shao \at
              University of Electronic Science and Technology of China \\
           \and
           Jingkuan Song$^*$, Corresponding author \at
              University of Trento, Italy \\
              \email{jingkuan.song@unitn.it} 
\and
           Xinyi Liu \at
              Qinzhou University, Guangxi, China
\and              
              Jiajun Liu \at
              Renmin University, Beijing, China
}

\date{Received: date / Accepted: date}

\maketitle
\begin{abstract}
During the last decade, the deluge of multimedia data has impacted a wide range of research areas, including multimedia retrieval, 3D tracking, database management, data mining, machine learning, social media analysis, medical imaging, and so on. 
Machine learning is largely involved in multimedia applications of building models for classification and regression tasks etc., and the learning principle consists in designing the models based on the information contained in the multimedia dataset.
While many paradigms exist and are widely used in the context of machine learning, most of them suffer from the `curse of dimensionality', which means that some strange phenomena appears when data are represented in a high-dimensional space. 
Given the high dimensionality and the high complexity of multimedia data, it is important to investigate new machine learning algorithms to facilitate multimedia data analysis.
To deal with the impact of high dimensionality, an intuitive way is to reduce the dimensionality. On the other hand, some researchers devoted themselves to designing some effective learning schemes for high-dimensional data.
In this survey, we cover feature transformation, feature selection and feature encoding, three approaches fighting the consequences of the curse of dimensionality. Next, we briefly introduce some recent progress of effective learning algorithms. Finally, promising future trends on multimedia learning are envisaged.

\keywords{High Dimensional \and Multimedia Data \and Learning \and Survery}

\end{abstract}

\section{Introduction}
\label{intro}

Today, large collections of digital multimedia data are continuously created in different fields and in many application contexts \cite{KarpathyCVPR14,Feng_2013_ICCV,ZhuHYSXL13,ZhuHSCX12,ZhuSS14b}. Multimedia finds its applications in various domains including, but not limited to, advertisements, journalism, cultural heritage, animal ecology, Web searching, geographic information systems, ecosystem, surveillance systems, entertainment, medicine, business and social services \cite{Choi_2015_CVPR,DBLP:journals/sigmod/KantorskiMH15,DBLP:conf/sigmod/2015,GaoSSZS15,zhu2015canonical,zhu2015subspace}. The vast amounts of new multimedia data in a large variety of formats (e.g., videos and images) and media modalities (e.g., the combination of, say, text, image, video and sound) are made available worldwide on a daily basis. Moreover, the quantity, complexity, diversity, high dimensionality and multi-modality of these multimedia data are all exponentially growing.

High-dimensional data pose many intrinsic challenges for pattern recognition problems \cite{icml11_ngiam,icml11_yao,scenecnn_iclr15,Gao_2015_CVPR,DBLP:conf/sigmod/2015}. For example, the curse of dimensionality and the diminishment of specificity in similarities between points in a high dimensional space~\cite{JMLR:v14:escalante13a,choi_pami13,CVPR14_Khosla}. Specifically, the complexity of many existing data mining algorithms is exponential with respect to the number of dimensions. With increasing dimensionality, existing algorithms soon become computationally intractable and therefore inapplicable in many real applications. 

An intuitive way is to reduce the number of input variables before a machine learning algorithm can be successfully applied \cite{Domingos:2012:FUT:2347736.2347755,DBLP:conf/bmvc/ChatfieldLVZ11,Wang6751553,kantorov2014}. The dimensionality reduction can be made in three different ways to support efficient search and effective storage: 1) feature transformation, which transforms existing high-dimensional features to a new reduced set of features by exploiting the redundancy, noisy or irrelevant of the input data and finding a smaller set of new variables, each being a combination of the input variables, containing basically the same information as the input variables; 2) feature selection, which selects a subset of the existing high-dimensional features by only keeping the most relevant variables from the original dataset; and 3) feature encoding, which  encodes the high-dimensional data into a compact code. Alternatively, some machine learning researchers are focusing on the design of efficient machine learning algorithms which can be directly applied to high-dimensional multimedia data.

The organization of the paper is given as follows. Section 2 presents a general framework of learning in high-dimensional multimedia data.  Sections 3-5 present some research works on dimensionality reduction, namely feature transformation, feature selection and feature encoding respectively. Section 6 briefly reviews some research efforts on efficient learning schemes for high-dimensional data. Finally, Section 7 provides some promising future trends and concludes this survey.

\section{A General Framework for Learning in High-Dimensional Multimedia Data}
\label{sec.mmIndex}

In this article, we refer to learning in high-dimensional data as: \textit{the problem of preprocessing a database of multimedia objects to provide low dimensional data for conventional machine learning algorithms or designing effective machine learning schemes for high dimensional multimedia features}.

\begin{figure}[t]
\centering
  \includegraphics[width=1\linewidth]{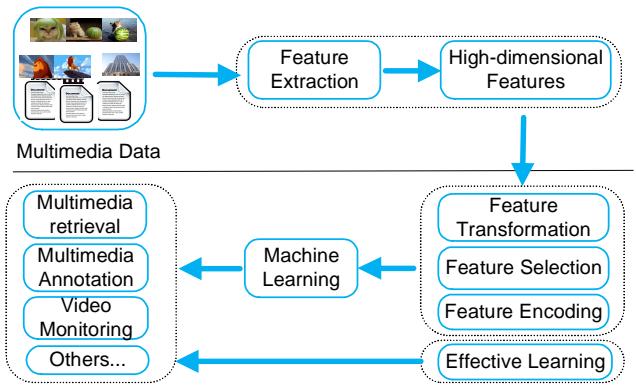}\\
  \caption{General framework for learning in high-dimensional multimedia data}
  \label{fig:framework}
\end{figure}

The general framework of learning in high-dimensional multimedia data is depicted in Fig.~\ref{fig:framework}. Firstly, high-dimensional multimedia features are extracted by using some common feature extraction techniques such as Histogram of Oriented Gradients (HOG), Speeded Up Robust Features (SURF), Local Binary Patterns (LBP), Haar wavelets, and color histograms. After the high-dimensional features are obtained, dimensional reduction techniques are often applied as a data pre-processing step to simplify the data model for computational efficiency and for improving the accuracy of the analysis. The techniques that can be employed for dimension reduction can be partitioned into three categories: 1) feature transformation; 2) feature selection; and 3) feature encoding. The outputs of feature reduction approaches are taken as the inputs for supervised, semi-supervised or unsupervised learning to support multimedia real applications such as multimedia retrieval, multimedia annotation and video tracking \cite{Gao_2015_CVPR,Wang_2015_CVPR,Shi_2015_CVPR,Hong_2015_CVPR}. Instead of reducing the high-dimensional multimedia data to fit traditional machine learning algorithms, some experts proposed effective learning schemas directly applied to these high-dimensional data to conduct advanced multimedia applications\cite{Nguyen_2015_CVPR,Papandreou_2015_CVPR,Zhang_2015_CVPR}.
In the following sections, we will discuss these techniques in details.

\section{Feature Transformation}
Feature transformation is a set of pre-processing techniques that aim to transform the original high-dimensional features to an alternative new set of features (predictor variables), while retaining as much information as possible by dropping less descriptive features from consideration. It has been widely researched in different fields including statistics and machine learning, with applications on object recognition \cite{DBLP:conf/eccv/GuptaGAM14,DBLP:journals/ijcv/GuptaAGM15,ECCV/2012/Ali}, data analysis and visualizations \cite{Saini:2012:MOM:2393347.2393373}, and many others \cite{Maaten2009,DBLP:conf/vluds/EngelHH11}.

Following \cite{DBLP:conf/vluds/EngelHH11}, feature transformation can be generally formulated as below. Let $ {X} \in \mathbb{R}^{m \times n}$, a set of $n$ data points in an $m$-dimensional feature space, and two 
distance (or dissimilarity) functions, ${d_m}:\mathbb{R}^m \times \mathbb{R}^m \to \mathbb{R}$ and ${d_t:\mathbb{R}^t \times \mathbb{R}^t \to \mathbb{R}}$, 
over the $m$-dimensional data space  and the target $t$-dimensional subspace respectively, with $ t \ll m$. A mapping function $\phi$ that maps the $m$-dimensional data points $({x}_i \in {X})$ to the $t$-dimensional target points $({y}_i \in {Y})$, i.e.,
\begin{align}
\phi : \mathbb{R}^m \to \mathbb{R}^t, 
{x}_i \to {y}_i, 1 \le  i \le n
\end{align}
is defined s.t. $\phi$ “faithfully” approximates pairwise distance relationships of ${X}$ by those of ${Y} \in \mathbb{R}^{t \times n}$, thereby mapping close (similar) points in data space to equally close points in target space, i.e., $d_m ({x}_i, {x}_j) \approx d_t ({y}_i, {y}_j)$, for $1 \le i, j \le n$. In particular, an adequate mapping is designed to ensure that remote data points are mapped to remote target points. Since the target space usually has lower degrees of freedom than those required to model distance relationships in the original space, the mapping $\phi$ adheres to an inherent error that is to be minimized by its definition. Thereby, $\phi$ is commonly defined to minimize the least squares error:
\begin{align}
{\varepsilon _\phi } \!=\! \sum\nolimits_{{1} \le i,j \le n} {{{M}_{i,j}}{{\left( {{d_m}\left( {{\mathbf{x}_i},{{x}_j}} \right) \!-\! {d_t}\left( {{{y}_i},{{y}_j}} \right)} \right)}^2}},  {M} \!\in\! \mathbb{R}^{n \times n} \nonumber
\label{eq.DimRed}
\end{align}
 where ${M}$ is a matrix used to define the weights of certain data relationships or dimensions.
Beside pairwise distance relationship preservation, there are also some variants on $\phi$ which are designed to preserve other relationships, such as the nearest neighbors relationship, or to minimize the errors measured by other factors depending on the distance functions used.

Feature transformation can be roughly categorized into linear transformation and non-linear transformation. For linear transformation, an explicit linear transformation function is learned to reduce the dimensionality and increase the robustness and the performance of domain applications. 
As one of the first dimension reduction techniques discussed in the literature, Principal Components Analysis (PCA)~\cite{jolliffe2002principal} conveys distance relationships of the data by orthogonally projecting it on a linear subspace of target dimensionality. In this specific subspace, the orthogonally projected data has maximal variance.
Latent Semantic Analysis (LSA) is a variant on the PCA idea presented in~\cite{DeerwesterDLFH90} and it has been employed on documents, images, videos and musics~\cite{He03}. 
Linear discriminant analysis (LDA)~\cite{hastie01statisticallearning} is a supervised subspace learning method which is based on Fisher Criterion. It aims to find a linear transformation $W\in\mathbb{R}^{d\times m}$ that maps $x_i$ in the d-dimensional space to a m-dimensional space, in which the between class scatter is maximized while the within-class scatter is minimized.
LDR~\cite{cunningham2015ldr} interprets linear dimensionality reduction in a simple optimization framework as a program with a problem-specific objective over orthogonal or unconstrained matrices. This framework gives insight to some rarely discussed shortcomings of well-known methods and further allows straightforward generalizations and novel variants of classical methods.

Linear transformation would be considered as a shortcoming in many applications and lots of research efforts have been devoted to non-linear feature transformation.
Multidimensional Scaling (MDS)~\cite{LafonL06}, also known as classical MDS, is a well-established approach that uses projection to map high-dimensional points to a linear subspace of lower dimensionality. The technique is often motivated by its goal to preserve pairwise distances in this mapping. As such, MDS defines a faithful approximation as one that captures pairwise distance relationships in an optimal way; more precisely, inner product relations. 
MDS has proven to be successful in many applications, but it suffers from the fact that it is based on Euclidean distances, and does not take into account the distribution of the neighboring datapoints. Isomap~\cite{TengLFCS05} is a technique that resolves this problem by attempting to preserve pairwise geodesic (or curvilinear) distances between datapoints. Geodesic distance is the distance between two points measured over the manifold. In Isomap, the geodesic distances between the datapoints $x_i~ (i \in \{1, 2,...,n\})$ are computed by constructing a neighborhood graph $G$, in which every datapoint $x_i$ is connected with its $k$ nearest neighbors. The shortest path between two points in the graph forms a good estimate of the geodesic distance between these two points, and can easily be computed using shortest-path algorithms. The low-dimensional representations $y_i$ of the datapoints $x_i$ in the low-dimensional space $Y$ are computed by applying multidimensional scaling on the resulting distance matrix.
In~\cite{Wichterich:2008:EES:1376616.1376639}, a method designed for Earth Mover’s Distance (EMD) is proposed to increase efficiency in the search process. It incorporates EMD assignment analysis among dimensions to obtain effective reduction. A tight EMD bound in the subspace is established to generate only a small number of candidates for the expensive EMD computations in the original space.  
Local Linear Embedding (LLE)~\cite{saul06spectral} is a local technique for dimensionality reduction that is similar to Isomap in that it constructs a graph representation of the datapoints. In contrast to Isomap, it attempts to preserve solely local properties of the data. \eat{As a result, LLE is less sensitive to short-circuiting than Isomap, because only a small number of properties are affected if short-circuiting occurs. Furthermore, the preservation of local properties allows for successful embedding of nonconvex manifolds. In LLE, the local properties of the data manifold are constructed by writing the datapoints as a linear combination of their nearest neighbors. In the low-dimensional representation of the data, LLE attempts to retain the reconstruction weights in the linear combinations as good as possible.}
LLE describes the local properties of the manifold around a datapoint $x_i$ by writing the datapoint as a linear combination $W_i$ (the so-called reconstruction weights) of its k nearest neighbors $x_{ij}$. Hence, LLE fits a hyperplane through the datapoint $x_i$ and its nearest neighbors, thereby assuming that the manifold is locally linear. Based on this weighted matrix, the low-dimensional space $Y$ can be computed. Some other non-linear feature transformation methods include kernel PCA~\cite{0026002}, Hessian LLE~\cite{citeulike:905977}, Laplacian Eigenmaps \cite{BelkinN03}, LTSA~\cite{ZhangZ04} and so on.

More recently, deep learning~\cite{LeeEN07,WangHWW14,RifaiVMGB11} have achieved great success for dimensionality reduction via the powerful representability of neural networks. A few algorithms that work well for this purpose, beginning with restricted Boltzmann machines (RBMs)~\cite{hinton2006reducing} and autoencoders \cite{BengioLPL06}.
In GAE~\cite{WangHWW14}, they extend the traditional autoencoder and propose a dimensionality reduction method by manifold learning, which iteratively explores data relation and use the relation to pursue the manifold structure. In CAE~\cite{RifaiVMGB11}, they add a well chosen penalty term to the classical reconstruction cost function and achieve results that equal or surpass those attained by other regularized autoencoders as well as denoising auto-encoders~\cite{VincentLLBM10} on a range of datasets.

\section{Feature Selection}
Feature selection methods provide us a way of reducing computation time, improving prediction performance, and a better understanding of the data in machine learning \cite{He6247966,icml2014c2_jawanpuria14,cvpr2013bbprbm}.
The focus of feature selection is to select a subset of variables from the input which can efficiently describe the input data while reducing effects from noise or irrelevant variables and still provide good prediction results~\cite{GuyonE03,icml2013pgbm,icassp2013deepemotion}. 
It can be broadly classified into two categories: 1) filter methods, which aim to remove irrelevant features from the original high-dimensional features or identify the most relevant subset features for maximally describing the information of the original information; and 2) wrapper methods, which choose a set of relevant features by searching through the feature space and then selecting the candidate feature subsets for supporting the highest predictor performance.

\subsection{Filter Methods}
Filter methods use variable ranking techniques as the principle criteria for variable selection by ordering \cite{Kourosh2008}. 
The most popular filter strategies for feature selection include Information gain~\cite{Reunanen03}, Chi-square measure~\cite{wu2002feature}, the Laplacian score (LS) \cite{He03}, odds ratio \cite{Mladenic98} and its derivatives~\cite{HeCN05}.
Ranking methods are used due to their simplicity and good performance is reported for practical applications. A suitable ranking criterion is used to score the variables and a threshold is used to eliminate variables below the threshold. Ranking methods are filter methods since they are applied before classification to filter out the less relevant variables. A basic property of a good feature is to contain useful information about the different classes in the data. This property can be defined as the feature relevance which provides a measurement of the discrimination power of a feature to different classes \cite{Kohavi97wrappersfor}.
Several publications \cite{HeCN05,Reunanen03} have presented various definitions and measurements for the relevance of a variable.

Next we look into two representative ranking methods. The input data $[x_{ij}, y_i]$ consists of $n$ samples $i= 1$ to $n$ with $m$ variables $j = 1$ to $m$, $x_i$ is the $i$th sample and $y_i$ is the class label for $x_i$.

A widely used criteria is the correlation criteria, which can be defined as:
\begin{align}
R\left( i \right) = {{{\mathop{ cov}} \left( {{x_i},Y} \right)}}/{{\sqrt {{\mathop{ var}} \left( {{x_i}} \right) \times {\mathop{ var}} \left( Y \right)} }}
\end{align}
where $x_i$ is the $i$-th variable, $Y$ is the output (class labels), $cov()$ is the covariance and $var()$ the variance. Note that correlation ranking can only detect linear dependencies between variable and target.

Information theoretic ranking criteria uses the measure of dependency (mutual information) between two variables. It is based on the observation that if one variable can provide information about the other then they are dependent. We start with Shannons definition for entropy given by:
\begin{align}
H\left( Y \right) =  - \sum\nolimits_y {p\left( y \right)\log \left( {p\left( y \right)} \right)}
\end{align}
It represents the uncertainty (information content) in the output $Y$. Suppose we observe a variable $X$ then the conditional entropy is given by:
\begin{equation}
H\left( {Y|X} \right) =  - \sum\nolimits_{x,y} {p\left( {x,y} \right)\log \left( {p\left( {y|x} \right)} \right)} 
\end{equation}
It implies that by observing a variable X, the uncertainty in the output Y. The decrease in uncertainty is given as:
\begin{align}
I\left( {Y,X} \right) = H\left( Y \right) - H\left( {Y|X} \right)
\end{align}
This gives the mutual information between $Y$ and $X$ meaning that if $X$ and $Y$ are independent then mutual information will be zero and greater than zero if they are dependent.
The definitions provided above are given for discrete variables and the same can be obtained for continuous variables by replacing the summations with integrations.

Recently, lots of research efforts have be devoted to filter-based feature selection. In~\cite{JavedBS12,RehmanJBS15} the authors develop a ranking criteria based on class densities for binary data. A two stage algorithm utilizing a less expensive filter method to rank the features and an expensive wrapper method to further eliminate irrelevant variables is used.
The advantages of feature ranking are that it is computationally light and it avoids overfitting and is proven to work well for certain datasets~\cite{LazarTMSCMSDBN12,GuyonE03}. Filter methods do not rely on learning algorithms which are biased. This is equivalent to changing data to fit the learning algorithm. One of the drawbacks of ranking methods is that the selected subset might not be optimal in that a redundant subset might be obtained. Some ranking methods such as Pearson correlation criteria  and MI do not discriminate the variables in terms of the correlation to other variables. The variables in the subset can be highly correlated in that a smaller subset would suffice~\cite{Chandrashekar201416}. In feature ranking, important features that are less informative on their own but  are informative when combined with others could be discarded~\cite{GuyonE03}. Finding a suitable learning algorithm can also become hard since the underlying learning algorithm is ignored. 

\subsection{Wrapper Methods}
Wrapper methods use the predictor as a black box and the predictor performance as the objective function to evaluate the variable subset. Since evaluating 2N subsets becomes a NP-hard problem, suboptimal subsets are found by employing search algorithms which find a subset heuristically. A number of search algorithms can be used to find a subset of variables which maximizes the objective function.
We broadly classify the wrapper methods into Sequential Selection Algorithms (SSA) and Heuristic Search Algorithms (HSA). The SSA start with an empty set and add features until the maximum objective function is obtained, while the HSA evaluate different subsets to optimize the objective function.
\eat{\begin{itemize}
\item . To speed up the selection, a criteria is chosen which incrementally increases the objective function until the maximum is reached with  minimum features. 
\item The HSA evaluate different subsets to optimize the objective function. Different subsets are generated either by searching around in a search space or by generating solutions to the optimization problem. 
\end{itemize}}

For SSA, Two representative methods are Sequential Feature Selection (SFS) and Sequential Floating Forward Selection (SFFS). The SFS algorithm \cite{Pudil:1994} starts with an empty set and adds one feature for the first step which gives the highest value for the objective function. Next, the remaining features are added individually to the current subset and the new subset is evaluated. While SFFS~\cite{Reunanen03} algorithm is more flexible than the naive SFS because it introduces an additional backtracking step. Specifically, the first step of the SFFS is the same as the SFS. Next, the SFFS excludes one feature at a time from the subset obtained in the first step and evaluates the new subsets. If excluding a feature increases the value of the objective function then that feature is removed and goes back to the first step with the new reduced subset or else the algorithm is repeated from the top. This process is repeated until the required number of features are added or the required performance is reached.

\eat{However, both SFS and SFFS methods suffer from producing nested subsets since the forward inclusion was always unconditional which means that two highly correlated variables might be included if it gave the highest performance in the SFS evaluation. To avoid the nesting effect, an Adaptive version of SFFS (ASFFS)~\cite{SomolPNP99} was developed and it  used a parameter $r$ which would specify the number of features to be added in the inclusion phase which was calculated adaptively. The parameter $o$ would be used in the exclusion phase to remove maximum number of features if it increased the performance. The ASFFS attempted to obtain a less redundant subset than the SFFS algorithm. It can be noted that a statistical distance measure can also be used as the objective function for the search algorithms as done in \cite{Pudil:1994,SomolPNP99}. Theoretically, the ASFFS should produce a better subset than SFFS but this is dependent on the objective function and the distribution of the data. The Plus-$L$-Minus-$r$ search \cite{Pudil:1994,NakariyakulC09} also tries to avoid nesting. In the Plus-$L$-Minus-$r$ search, in each cycle $L$ variables were added and $r$ variables were removed until the desired subset was achieved. The parameters $L$ and $r$ have to be chosen arbitrarily. In \cite{NakariyakulC09} the authors tried to improve the SFFS algorithm by adding an extra step after the backtracking step in the normal SFFS in which a weak feature is replaced with a new better feature to form the current subset. }

For HSA, Genetic Algorithm (GA)~\cite{Goldberg:1989:GAS:534133} can be used to find a subset of features where in the chromosome bits represent if the feature is included or not. The global maximum for the objective function can be found which gives the best suboptimal subset. Here again the objective function is the predictor performance. The GA parameters and operators can be modified within the general idea of an evolutionary algorithm to suit the data or the application to obtain the best performance or the best search results. Multi-objective GA is used for hand written digit recognition in \cite{OliveiraSBS03}. 

\eat{Several wrapper methods are compared with different datasets in \cite{KudoS00}. They derive various fitness functions with weighting imposing characteristics. A binary PSO \cite{TuCCY06} algorithm can also be used for wrapper implementation.
The main drawback of wrapper methods is the number of computations required to obtain the feature subset. For each subset evaluation, the predictor creates a new model i.e. the predictor is trained for each subset and tested to obtain the classifier accuracy. If the number of samples is large, most of the algorithm execution is spent in training the predictor. Another drawback of using the classifier performance as the objective function is that the classifiers are prone to overfitting \cite{Kohavi97wrappersfor}. Overfitting occurs if the classifier model learns the data too well and provides poor generalization capability.Using classification accuracy in subset selection can result in a bad feature subset with high accuracy but poor generalization power. To avoid this, a separate holdout test set can be used to guide the prediction accuracy of the search~\cite{Kohavi97wrappersfor}.}

\section{Feature Encoding}

Feature encoding methods encode the high-dimensional data into compact codes so that efficient retrieval and effective storing can be conducted.
\subsection{Quantization}
The general quantization problem can be formulated as $\zeta : {x} \in \mathbb{R}^m \longrightarrow \{0,1,...,2^L\}$, where $\zeta$ is the function to quantize an $m$-dimensional feature vector $x$ to a value in $\{0,1,...,2^L\}$ and $L$ is the number of bits for approximating the $m$-dimensional feature vectors. For the special case of $m$=1 (called scalar quantization), a scalar input is provided, and it implies that a range of scalar quantities are quantized into a single integer (or the same bit-string). For the cases of $m > 1$ (called vector quantization), it means that a group of vectors are approximated into the same bit-string. 

\subsubsection{Scalar Quantization}

Scalar quantization takes a real-valued scalar quantity and maps it into one of a finite set of values.  The idea of using quantization for high-dimensional indexing to overcome the ``dimensionality curse" first appeared in \cite{DBLP:conf/vldb/WeberSB98}, where a very simple scalar quantization scheme called Vector-Approximation file (VA-file) is proposed. 
For each dimension $i$, the VA-file partitions the one-dimensional data range into $2^{L_i}$ slices where $L_i$ is the number of bits for representing the dimension. Each dimension can then be approximated by $L_i$ bits by checking to which slice the value on dimension $i$ belongs (i.e., map the value into one of 0 to $2^{L_i}$-1). Given an $m$-dimensional feature vector, a bit-string of length $L=\sum L_i$ by concatenating all its dimensions' bits is used to approximate the original feature vector. 
\eat{The VA-file is basically an array of bit-strings which are compact and geometric approximations of the original data. Similarity search is conducted by accessing the entire VA-file to exclude most vectors from being accessed (i.e., the filtering step) based on the approximated distances from bit-strings. }

\subsubsection{Vector Quantization}

Vector quantization works by dividing all the feature vectors into groups (or clusters). It takes a feature vector and then maps it into one of the finite set of groups, where each group is approximated by its centroid point. Depending on the number of groups needed, different numbers of bits $b$ can be determined to encode the group IDs. An example of vector quantization is the $K$-means clustering. Formally, the standard $K$-means clustering is defined as below \cite{Lloyd82}.

Given $N$ feature vectors ${X} = \{{x}_1, \cdots, {{x}}_n\}\in \mathbb{R}^{m\times n}$, the $K$-means algorithm partitions the database into $K$ clusters, each of which associates one codeword ${{c}}_i\in \mathbb{R}^{m\times 1}$. Let ${{C}} = [{{c}}_1, \cdots, {{c}}_K]\in \mathbb{R}^{m\times K}$ be the corresponding codebook. 
Then the codebook is learned by minimizing the within-cluster distortion, e.g.,
\begin{eqnarray}
\min~~&{\sum_{i}{\|{{x}}_i - {{C}}{{v}}_i\|_2^2}} \\\nonumber
\operatorname{s.t.}~~&{{v}}_i \in \{0, 1\}^{K\times 1}, \forall i , \|{{v}}_i\|_1 = 1, \forall i
\end{eqnarray}
where ${{v}}_i$ is a $1$-of-$K$ encoding vector (i.e., $K$ dimensions with one $1$ and $(K-1)$ $0$s) 
to indicate which \textit{codeword} is used to map ${{x}}_i$, and $\|\cdot\|_1$ is the $\ell_1$ norm. 

In vector quantization, vectors are quantized into clusters. Given a query vector, it is firstly mapped into the closest clusters, followed by computing the distances between the query vector and all the feature vectors inside those clusters. For large-scale high-dimensional databases, it is very challenging to determine the value of $K$. When $K$ is too small, coarse quantization is performed, i.e., too many vectors are approximated into the same cluster, leading to many database vectors to be accessed and compared. Although a larger $K$ value results in finer quantization, consequently more clusters need to be accessed to search the closest ones. To maintain high-quality quantization, large $K$ values are usually required. To index a large number of clusters, they can be hierarchically organized in a vocabulary tree which directly defines the vector quantization \cite{DBLP:conf/cvpr/NisterS06}. The quantization and the indexing can then be integrated. Naturally, it also inherits the drawbacks of tree structures to a certain degree.

 \subsubsection{Product Quantization}
 Scalar quantization quantizes each dimension of the vector and may over-quantize the data since each dimension may require multiple bits. On the contrary, vector quantization quantizes the full-dimensional vectors as a whole and may under-quantize the data since each vector needs a few bits only, independent of the dimensionality. Scalar quantization suffers from scanning relatively large-sized approximations, while vector quantization has the scalablity issue when comparing a large number of clusters. 
 To address these problems, product quantization (PQ) \cite{DBLP:journals/pami/JegouDS11} is proposed to perform quantization on subvectors of the original full-dimensional vectors, i.e., an intermediate of scalar quantization and vector quantization. 
 
 The key idea of PQ is to split an $m$-dimensional vector into $P$ disjoint subvectors on which quantization is then performed.
 Assume the $j$-th subvector contains $m_j$ dimensions and then $m=\sum_{j = 1}^{P}{m_j}$. Without loss of generality, $m_j$ is set to $ m/P$ and $m$ is assumed to be divisible by $P$.
 For each subvector, $K$-means is performed on all the database vectors to obtain $K$ \textit{sub codewords}. 
 With $K\times P$ clusters generated from $P$ subvectors, it can have the capacity to represent 
 $K^P$ possible clusters on the original $m$-dimensional vector space based on the \textit{Cartesian product} relationship among subvectors. 
  
  Let ${{C}}^j \in \mathbb{R}^{m_j \times K}$ be the matrix of the $j$-th sub codebook and each column is an $m_j$-dimensional sub codeword.
 As discussed in~\cite{NorouziF13,WangWSXSL15}, PQ can be taken as optimizing the following problem with respect to ${{C}}^j$ and ${{v}}_i^j$, where ${{v}}_i^j$ is the 1-of-$K$ encoding vector on the $j$-th subvector and the index of $1$ indicates which sub codeword is used to encode the $j$-th subvector for the $i$-th vector ${{x}}_i$: 
 \begin{align}
 \begin{split}
 \min ~~ f_{\text{PQ}, P, K} = \sum_{i}
 {
 	\left\|
 		{{x}}_i -
 			\begin{bmatrix}
 			{{C}}^1 {{v}}_i^1\\
 			\vdots \\
 			{{C}}^{P} {{v}}_i^P
 		\end{bmatrix}
 	\right\|_2^2
 } \\
 \operatorname{s.t.}~~{{v}}_i^{j} \in \{0, 1\}^{K\times 1}, \forall i, j,
  \|{{v}}_i^{j}\|_1 = 1, \forall i,j.
 \end{split}
 \label{eqn:pq}
 \end{align}

The main advantage of PQ lies in the fact that only a very small number of clusters generated from subvectors are used to approximate the full Euclidean distance. Therefore, the required memory cost is small. To avoid an exhaustive scan on the database codes, an inverted file is also combined to index clusters. Given a query, the inverted file is first accessed to limit the search space, then more accurate Euclidean distances can be computed based on the subvector-to-centroid distances.

\eat{For quantization methods, the bit-string approximations are much smaller in space than the original data. The approximations can be quickly scanned to approximate the distances, where auxiliary data structures can also be used to help (e.g., tree structures can be combined with the VA-file and PQ requires cluster lookup tables). They are generally robust to feature vector dimensionality, thus overcome the `dimensionality curse' and outperform tree structures in general. However, quantization methods incur a large number of float-valued distance approximations based on expensive measures such as the Euclidean distance. Efficient distance approximation is strongly desired. } 

\subsection{Hashing}
Very recently, hashing~\cite{ZhuHSZ13,ZhuZH14,ZhuHCCS13,song2013effective,ZouCSZYS15,MM15zou,MM15song,MM15gao,ZouLLFYL13,ZouFLLYLL13} has attracted enormous attention from different research areas to achieve fast similarity search due to its high efficiency in terms of the storage and computational cost. The basic idea of hashing is to encode a high-dimensional data point (or feature vector) into a binary code (i.e., a bit-string). Different from quantization, for two binary codes in hashing, their Hamming distance in the Hamming space can be directly employed to measure the closeness between two corresponding high-dimensional feature vectors. Compared with distances computed on real-valued high-dimensional feature vectors, the Hamming distance computed on binary codes is far more efficient by using bit operations. The form of hashing can be formulated as below: 
$
\mathbf{y} = {\mathcal{H}}({x}) \in \{0, 1\}^{L},
$
where ${{x}}$ is an $m$-dimensional real-valued vector, $\mathbf{y}$ is the corresponding binary code for ${x}$ with $L$ bits, and $\mathcal{H}$ is the hash function family to map ${x}$ into $\mathbf{y}$. Typically, in the hash function family, one hash function is used to generate one bit of the binary code, i.e., $ \mathcal{H}({x}) = ({h}_1({x}),\cdots,{h}_L({x})).$

The most challenging problem on hashing is how to generate effective hash functions which can perform approximate similarity search as accurately as possible. In other words, hashing functions should preserve the similarity relationship among feature vectors from the original high-dimensional feature space to the mapped Hamming space. 
According to the characteristics of training data, they can be summarized into unsupervised hashing, supervised hashing, and semi-supervised hashing.

For unsupervised hashing where training data have no labels, many works exploit the low energy spectrum of data neighborhood graphs to obtain the hash codes and hash functions \cite{Weiss08,DBLP:conf/icml/LiuWKC11}. This group of methods usually requires building a neighborhood graph and computing eigenvalues of this graph Laplacian. An alternative solution is to seek principled linear projections using PCA \cite{DBLP:journals/pami/WangKC12} or sparse coding \cite{ZhuHCCS13}.

While unsupervised hashing shows promise in retrieving neighbors based on metric distances, e.g., $\ell_2$ distance, a variety of practical applications prefer semantically similar neighbors \cite{SHCVPR2008}. Therefore, some recent works have also incorporated supervised information to improve the hashing performance. Such supervised information can be customarily expressed as labels of similar (or relevant) and dissimilar (or irrelevant) data pairs when available. Typical supervised hashing methods include Semantic Hashing \cite{salakhutdinov2007semantic}, Minimal Loss Hashing (MLH) \cite{DBLP:conf/icml/NorouziF11}, Linear Discriminant Analysis Hashing (LDAHash) \cite{strecha2012ldahash}, Kernelized Supervised Hashing (KSH) \cite{DBLP:conf/cvpr/LiuWJJC12}, Order Preserving Hashing (OPH) \cite{Wangjianfeng13}, etc. Supervised hashing is expected to achieve better result quality than unsupervised hashing, if the supervised information is properly utilized.

One of the main problems with supervised hashing methods is that very noisy or limited training data may easily lead them to be ineffective or over-fitting. To tackle this problem,
Semi-Supervised Hashing (SSH) has also been studied \cite{DBLP:journals/pami/WangKC12}. SSH aims to minimize the empirical errors on the labeled training data. At the same time, it also maximizes the variance and the independence of different bits on both the labeled data and the unlabeled data. Lai et al.\cite{Lai_2015_CVPR} proposed a "one-stage" supervised hashing method via a deep architecture that maps input images to binary codes. The proposed deep architecture uses a triplet ranking loss designed to preserve relative similarities. Semantics Preserving Hashing (SePH) \cite{Lin_2015_CVPR} 
method is proposed for cross-view retrieval. It utilizes kernel logistic regression with a sampling strategy as basic hash functions to model the projections from features in each view to the hash codes.  Hashing across Euclidean space and Riemannian
manifold (HER) is proposed by deriving a unified framework
to firstly embed the two spaces into corresponding reproducing
kernel Hilbert spaces, and then iteratively optimize
the intra- and inter-space Hamming distances in a maxmargin
framework to learn the hash functions for the two
spaces \cite{Lin_2015_CVPR}. Table~1 shows the comparative results of some standard hashing methods in precision on the ILSVRC 2012 dataset. 

\begin{table}[h]
\begin{center}
\caption{Comparative results in precision of Hamming distance on the ILSVRC 2012 datase(10,000 samples are used for training).}
\begin{tabular}{|c|c|c|}
\hline
 Method & 64 bits & 128 bits\\ \hline
\multirow{4}{*}{} 
      BRE \cite{NIPS2009_3667} & High & low\\
      SSH \cite{DBLP:journals/pami/WangKC12} & low & low\\
      ITQ \cite{DBLP:journals/pami/GongLGP13}  & fair & fair\\
      KSH \cite{DBLP:conf/cvpr/2012}  & High & fair\\
      FastHash \cite{DBLP:journals/corr/LinSSHS14}  & High & High\\
 \hline
  Dataset & Dim & Reference Set\\ \hline
 \multirow{4}{*}{} 
  MNIST & 784 & 60K\\
       SIFT10K & 128 & 10K\\
       SIFT1M & 128 & 1M\\
       GIST1M & 960 & 1M\\
       Tiny1M & 384 & 1M\\
       SIFT1B & 128 & 1B\\
 \hline       
\end{tabular}
\end{center}
\label{tab.res1}
\end{table}

There are also some further developments extended from the standard hashing algorithms:
Multiple Feature Hashing (MFH) \cite{DBLP:conf/mm/SongYHSH11} learns a group of hash functions to generate binary codes for multiple visual features generated from the same media type, e.g., videos.  
Cross-Media Hashing \cite{Kumar:2011:LHF:2283516.2283623,DBLP:conf/sigmod/SongYYHS13,DBLP:conf/kdd/ZhenY12} considers different media types of data from heterogeneous data sources (e.g., text documents from Google and images from Facebook) to support cross-media retrieval and jointly feeds them into the same model for efficient hashing. 
Complementary Hashing \cite{XuWLZLY11} employs multiple complementary hash tables to further improve the search quality. It balances the recall and the precision in a more effective way.  
Weighted Hamming Distance (WHD) \cite{zhang2013} has been proposed to rank the binary codes of hashing methods to improve the ranking quality based on the Hamming distance. Moreover, robust hashing with local modes (RHLM) \cite{6714849} for accurate and fast approximate similarity search have been proposed. Specifically, it uses the learned hash functions, and all the database points are then mapped into their binary hash codes and organized into buckets. Data points having the same hash code belong to a single bucket that is identified by the hash code. Given a query data point, approximate similarity search can be efficiently
achieved by exploring the buckets, which have similar hash
codes to the query hash code. A comparison with the state-of-the art single modality and multiple modalities hash learning methods is shown in \cite{Lin_2015_CVPR}.

\section{Effective Learning Schemes}
While dimensional reduction is a possible way to alleviate the effect of high-dimensionality, some effective learning schemes are proposed for online learning and paralleling computing to directly deal with the high-dimensional data.

Online learning~\cite{Shalev-Shwartz12} is a well established learning paradigm which has both theoretical and practical appeals. The key difference between online learning and batch learning techniques, is that in online learning the mapping is updated after the arrival of every new data point in a scalable fashion, while batch techniques are used when one has access to the entire training dataset at once. Therefore, online learning is able to deal with scalable and high-dimensional data. Online learning has been studied in several research fields including game theory, information theory and machine learning~\cite{mcmahan2011follow,BartlettHR07}.

Online learning is performed in a sequence of consecutive rounds,
where at round t the learner is given a question, $x_t$, taken from an
instance domain $X$ , and is required to provide an answer to this question,
which we denote by $p_t$. After predicting an answer, the correct
answer, $y_t$, taken from a target domain $Y$, is revealed and the learner
suffers a loss, $l(p_t,y_t)$, which measures the discrepancy between his
answer and the correct one. While in many cases $p_t$ is in $Y$, it is sometimes
convenient to allow the learner to pick a prediction from a larger set, which  we denote by $D$.
Many successful algorithms have been developed over the past few years to optimize the online learning problem, and they differ in the choice of loss function and regularization term. A modern view of these algorithms casts the problem as the task of following the regularized leader (FTRL)~\cite{mcmahan2011follow}. Informally, FTRL methods choose the best decision in hindsight at every iteration. Verbatim usage of the FTRL approach fails to achieve low regret, however, adding a proximal term to the past predictions leads to numerous low regret algorithms~\cite{HazanK10}. The proximal term strongly affects the performance of the learning algorithm. Therefore, adapting the proximal function to the characteristics of the problem at hand is desirable. 
Online gradient descent~\cite{BartlettHR07} generalizes this FTRL by deriving a simple reduction from convex functions to linear functions. In~\cite{DuchiHS10}, the authors present a new family of sub-gradient methods that dynamically incorporate knowledge of the geometry of the data observed in earlier iterations to perform more informative gradient based learning. Mirror descent~\cite{BartlettHR07} generalizes online gradient descent, and considers the best-experts problem, where on each round we must choose an `expert' from a set whose advice we follow for that round. 

The high-dimensional and big data phenomenon is intrinsically related to the paralleling computing. Large companies as Facebook, Yahoo!, Twitter, LinkedIn benefit and contribute working on paralleling computing projects. Big Data infrastructure deals with Hadoop, and other related software as: 1) Apache Hadoop, a software for data-intensive distributed applications, based in the MapReduce programming  model and a distributed file system called Hadoop Distributed Filesystem (HDFS). Hadoop allows writing applications that rapidly process large amounts of data in parallel on large clusters of compute nodes. A MapReduce job divides the input dataset into independent subsets that are processed by map tasks in parallel. This step of mapping is then followed by a step of reducing tasks. These reduce tasks use the output of the maps to obtain the final result of the job.  2) Apache S4~\cite{NeumeyerRNK10}: platform for processing continuous data streams. S4 is designed specifically for managing data streams. S4 apps are designed combining streams and processing elements in real time.
\section{Future Trends and Conclusion}

Hign-dimensionality and big data is going to continue growing during the next years, especially with the emergence of deep learning and social media. The data is going to be more diverse, larger, and complex. Therefore, dimension reduction algorithms are still going to be a hot research topic in the near future. On the other hand, effect learning algorithms for first-order optimization, online learning and paralleling computing will be more desired.

We discussed in this paper some insights about the learning in high-dimensional data. We first consider the main challenges for the high-dimensional data and then survey some existing techniques dealing with the problem. Finally, we conclude this paper by providing some possible future trends.

\section{Acknowledge}
The work of Lianli Gao has been partially supported by NSFC (Grant No.61502080) and by the Fundamental Research Funds for the Central University (Grant No.ZYGX2014J063). The work of Junming Shao has been supported partially by NSFC (Grant No. 61403062, 61433014), and Fundamental Research Funds for the Central Universities (Grant No. ZYGX2014J053).
{\small

}

\end{document}